\documentclass[journal]{IEEEtran}
\IEEEoverridecommandlockouts
\ifCLASSINFOpdf
\else
   \usepackage[dvips]{graphicx}
\fi
\usepackage{url}

\usepackage{cite}
\usepackage{amsmath,amssymb,amsfonts}
\usepackage[ruled,vlined]{algorithm2e}
\usepackage{graphicx}
\usepackage{textcomp}
\usepackage{xcolor}
\usepackage{float}
\usepackage{multicol}
\usepackage{mathtools}
\usepackage[export]{adjustbox}
\usepackage{url}
\def\etal{\emph{et al.}}   
\DeclareMathOperator*{\argmax}{arg\,max}
\DeclareMathOperator*{\argmin}{arg\,min}
\begin{document}

\title{Meta Generative Attack on Person Reidentification}

\author{A V Subramanyam, \IEEEmembership{Member, IEEE}
\thanks{A V Subramanyam is with IIITD, India (e-mail: subramanyam@iiitd.ac.in). We acknowledge the support of the IHUB-ANUBHUTI-IIITD FOUNDATION setup under the NM-ICPS scheme of the DST, India.}}

\markboth{Journal of \LaTeX\ Class Files, Vol. 14, No. 8, August 2015}
{Shell \MakeLowercase{\textit{et al.}}: Bare Demo of IEEEtran.cls for IEEE Journals}
\maketitle

\begin{abstract}
Adversarial attacks have been recently investigated in person re-identification. These attacks perform well under cross dataset or cross model setting. However, the challenges present in cross-dataset cross-model scenario does not allow these models to achieve similar accuracy. To this end, we propose our method with the goal of achieving better transferability against different models and across datasets. We generate a mask to obtain better performance across models and use meta learning to boost the generalizability in the challenging cross-dataset cross-model setting. Experiments on Market-1501, DukeMTMC-reID and MSMT-17 demonstrate favorable results compared to other attacks.
\end{abstract}

\begin{IEEEkeywords}
Adversarial attacks, Meta learning, ReID.
\end{IEEEkeywords}

\section{Introduction}
The tremendous performance of deep learning models has led to rampant application of these systems in practice. However, these models can be manipulated by introducing minor perturbations {\cite{szegedy2013intriguing, goodfellow2014explaining, wang2020you, wang2020adversarial, zhang2022local}}. This process is called adversarial attacks. In case of person re-identification, for a given query input $x$, a target model $f$ and a gallery, the attack is defined as,
\begin{align}
    &\lVert f(\mathbf{x}+\boldsymbol{\delta}) - f(\mathbf{x}_g)\rVert_2 > \lVert f(\mathbf{x}+\boldsymbol{\delta}) - f(\bar{\mathbf{x}}_g)\rVert_2 \;\;\;\textit{s.t.}\; \lVert \boldsymbol{\delta} \rVert_p \leq \epsilon, \nonumber\\
    &\mathbf{x}_g \ni topk(\mathbf{x}+\boldsymbol{\delta}), ID(\mathbf{x}) = ID(\mathbf{x}_g) \neq ID(\bar{\mathbf{x}}_g) \nonumber
\end{align}
where $\mathbf{x}_g$ and $\bar{\mathbf{x}}_g$ are gallery samples belonging to different identity and $\boldsymbol{\delta}$ is the adversarial perturbation with an $l_p$ norm bound of $\epsilon$. {\textit{topk}($\cdot$)} refers to the top $k$ retrieved images for the given argument.

 Adversarial attacks have been extensively investigated under classification setting \cite{akhtar2021advances} and also studied in other domains \cite{li2021concealed, li2021simple, jia20203d} in the recent times. However, {to the best of our knowledge}, there are very few works which study these attacks in person re-identification domain. In the following we briefly discuss some classical attacks under classification setting. Szegedy \etal~\cite{szegedy2013intriguing} proposed the first work on generation of adversarial sample for deep neural networks using L-BFGS. Goodfellow \etal~\cite{goodfellow2014explaining} proposed an efficient adversarial sample generation method using fast gradient sign method (FGSM). Kurakin \etal~\cite{kurakin2016adversarial} proposed an iterative FGSM method. Other prominent works include \cite{madry2017towards,carlini2017towards,papernot2016limitations,dong2018boosting,croce2020reliable,wang2021feature}.

In person re-id {\cite{zhou2019omni,chang2018multi,li2019cross,yang2021pixel}}, both white-box and black box attacks have been proposed in \cite{yang2021learning, ding2021beyond, wang2020transferable, li2021qair}. These attacks use a labeled source dataset and show that the attacks are transferable under cross-dataset or cross-model, or both settings. However, transferabilty of attacks in the challenging cross-dataset and cross-model setting is still an issue. In this work, we propose to use a mask and meta-learning for better transferability of attacks. We also investigate adversarial attacks in a completely new setting where the source dataset does not have any labels and the target model structure or parameters are unknown. 

\section{Related Works}

In \cite{9226484}, authors propose white box and black box attacks. The black box attack only assumes that the victim model is unknown but the dataset is available. \cite{wang2019advpattern} introduces physically realizable attacks in white box setting by generating adversarial clothing pattern. \cite{li2021qair} proposes a query based attack wherein the images obtained by querying the victim model are used to form triplets for triplet loss. \cite{bouniot2020vulnerability} proposes white box attack using self metric attack; wherein the positive sample is obtained by adding noise to the given input and obtain negative sample from other images. In \cite{yang2021learning}, authors propose a meta-learning framework using a labeled source and extra association dataset. This method generalizes well in cross-dataset scenario. In \cite{ding2021beyond}, Ding~\etal ~proposed to use a list-wise attack objective function along with model agnostic regularization for better transferability. A GAN based framework is proposed in \cite{wang2020transferable}. Here the authors generate adversarial noise and mask by training the network using triplet loss.

In this work we use a GAN network to generate adversarial sample. In order to achieve better transferability of attack across models, we suppress the pixels that generate large gradients. Suppressing these gradients allows the network to focus on other pixels. In this way, the network can focus on pixels that are not explicitly salient with respect to the model used for attack. We further use meta learning \cite{finn2017model} which also allows incorporation of an additional dataset to boost the transferability. We refer this attack as Meta Generative Attack (MeGA). Our work is closest in spirit to \cite{yang2021learning, wang2020transferable}, however, the mask generation and application of meta learning under GAN framework are quite distinct from these works.
%Our method is closest in spirit to \cite{wang2020transferable}. However, the mask generation process are completely different proposed setting of learning from unlabeled source dataset and .

\iffalse
\textbf{Adversarial Defense} Countering adversarial attacks, the goal of adversarial defense is to achieve the accuracy comparable to that of untargeted model. The defense methods either use adversarial examples during training or modify the network itself. Adversarial training is often considered as a first line of defense \cite{szegedy2013intriguing, goodfellow2014explaining, moosavi2016deepfool} and also demonstrates the strongest defense. Among other class of defenses which modify the network are defensive distillation \cite{papernot2016distillation}, gradient regularization \cite{ross2018improving}, biologically inspired models \cite{nayebi2017biologically, krotov2018dense}, convex ReLU relaxation \cite{wong2018provable}, image enhancement \cite{mustafa2019image}, image restoration \cite{zhao2021removing}. 
\fi

\section{Methodology}
In this work we address both white-box and black-box attacks. We need that the attack is transferable across models and datasets. Thus if we obtain the attack sample using a given model $f$, the attack is inherently tied to $f$ \cite{wang2021feature}. In order that attack does not over-learn, we apply a mask that can focus on regions that are not highly salient for discrimination. This way the network can focus on less salient but discriminative regions thereby increasing the generalizability of attack to other models. On the other hand, meta learning has been efficiently used in adversarial attacks \cite{yuan2021meta, yang2021learning, feng2021meta} to obtain better transferability across datasets. However meta learning has not been explored with generative learning for attacks in case of PRID. We adapt the MAML meta learning framework \cite{finn2017model} in our proposed method. While the black box attack works assume the presence of a labeled source dataset, we additionally present a more challenging setting wherein no labels are available during attack.
\begin{figure}
\centering
\includegraphics[width = .45\textwidth]{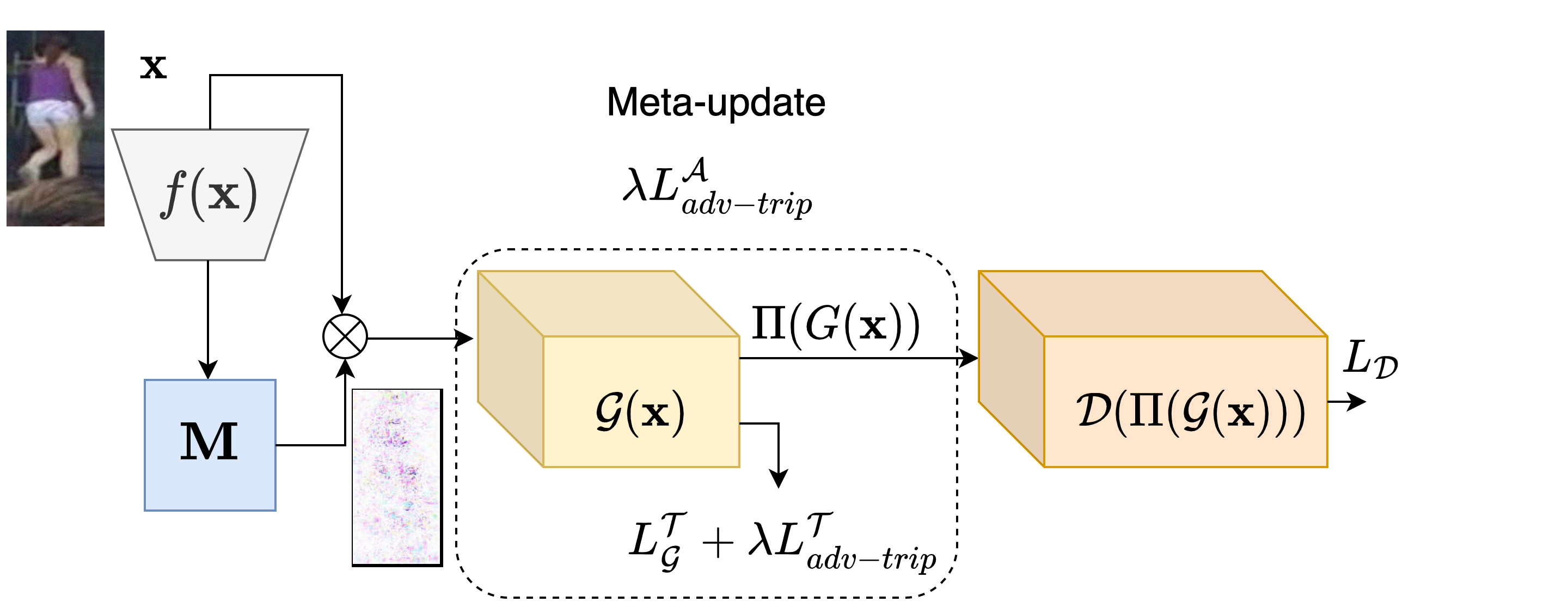}
\caption{Model architecture. Mask $\mathbf{M}$ is generated using model $f$ and is used to mask the input $\mathbf{x}$. GAN is trained using a meta learning framework with an adversarial triplet loss and GAN loss.}
\label{fig:architecture}
\end{figure}  

Our proposed model is illustrated in Figure \ref{fig:architecture}. In case of white-box setting, 
the generator $\mathcal{G}$ is trained using the generator loss, adversarial triplet loss and meta learning loss while the discriminator $\mathcal{D}$ is trained with the classical binary cross-entropy discriminator loss. The mask is obtained via self-supervised triplet loss. The network learns to generate adversarial image. While the GAN loss itself focuses on generating real samples, the adversarial triplet loss guides the network to generate samples that will be closer to negative samples and farther away from positive samples. 
\subsection{GAN training} 
Given a clean sample $\mathbf{x}$, we use the generator $\mathcal{G}$ to create the adversarial sample $\mathbf{x}_{adv}$. The overall GAN loss is given by, $\mathcal{L}_{GAN} = E_{\mathbf{x}}\log \mathcal{D}(\mathbf{x}) + E_{\mathbf{x}}\log(1 - \mathcal{D}(\Pi(\mathcal{G}(\mathbf{x}))))$.
% \begin{equation}
%     \mathcal{L}_{GAN} = E_{\mathbf{x}}\log \mathcal{D}(\mathbf{x}) + E_{\mathbf{x}}\log(1 - \mathcal{D}(\Pi(\mathcal{G}(\mathbf{x}))))
% \end{equation}
Here $\Pi(.)$ denotes the projection into $l_{\infty}$ ball of $\epsilon$-radius within $\mathbf{x}$ and $\mathbf{x}_{adv} = \Pi(\mathcal{G}(\mathbf{x}))$. In order to generate adversarial samples, a deep mis-ranking loss is used \cite{wang2020transferable},
\begin{align}
    \mathcal{L}_{adv-trip}(\mathbf{x}_{adv}^{a}, \mathbf{x}_{adv}^{n}, \mathbf{x}_{adv}^{p}) &= \max(\lVert \mathbf{x}_{adv}^{a} - \mathbf{x}_{adv}^n\rVert_2 \label{eq:adv-triplet} \\ \nonumber
    &- \lVert \mathbf{x}_{adv}^{a} - \mathbf{x}_{adv}^p\rVert_2 + m,0)
\end{align}
where $m$ is the margin. {$\mathbf{x}_{adv}^{a}$ is the adversarial sample obtained from anchor sample $\mathbf{x}^{a}$. Similarly, $\mathbf{x}_{adv}^{p}$ and $\mathbf{x}_{adv}^{n}$ are the adversarial samples obtained from respective positive and negative samples $\mathbf{x}^{p}$ and $\mathbf{x}^{n}$.} This loss tries to push the negatives closer to each other and pulls the positives farther away. Thus the network learns to generate convincing adversarial samples.
\subsection{Mask Generation}
Attack obtained using the given model $f$ leads to poor generelization to other networks. In order to have a better tranferability, we first compute the gradients with respect to self-supervised triplet loss $\mathcal{L}_{adv-trip}(\mathbf{x},\mathbf{x}^n,\mathbf{x}^p)$, where $\mathbf{x}^p$ is obtained by augmentation of $\mathbf{x}$ and $\mathbf{x}^n$ is the sample in the batch which lies at a maximum Euclidean distance from $\mathbf{x}$. Here, the large gradients are primarily responsible for loss convergence. Since this way of achieving convergence is clearly coupled with $f$, we mask the large gradients. Thus, the convergence is not entirely dependent on the large gradients and focuses on other smaller ones which can also potentially posses discriminative nature. Thus the overfitting can be reduced by using the mask. To obtain the mask, we compute,
\begin{equation}
    \mathbf{grad}_{adv-triplet} = \nabla_{\mathbf{x}}\mathcal{L}_{adv-trip}(\mathbf{x},\mathbf{x}^n,\mathbf{x}^p) 
    \label{eq:grad}
\end{equation}
Note that, we use the real samples in Eq. \ref{eq:grad}.
The mask is given by $\mathbf{M} = sigmoid(\lvert \mathbf{grad}_{adv-triplet} \rvert)$, where $\lvert \cdot \rvert$ denotes absolute value. We mask $\mathbf{x}$ before feeding as an input to the generator $\mathcal{G}$. The masked input is given as $\mathbf{x} = \mathbf{x}\odot (1-\mathbf{M})$, where $\odot$ denotes Hadamard product. 

{Masking techniques have also been explored in \cite{parascandolo2020learning, shahtalebi2021sand} where the idea is to learn the model such that it does not overfit to the training distribution. Our masking technique is motivated from the idea that an adversarial example should be transferbale across different reid models. Our technique is distinct and can be applied to an individual sample. Whereas, masking technique in \cite{parascandolo2020learning, shahtalebi2021sand} seeks agreement among the gradients obtained from all the samples of a batch.
This technique in \cite{parascandolo2020learning, shahtalebi2021sand} also suffers from the drawback of tuning hyperparameter. Further, the masking technique of \cite{parascandolo2020learning} is boolean while ours is continuous.}

\subsection{Meta Learning}
Meta optimization technique allows to learn from multiple datasets for different tasks while generalizing well on a given task. One of the popular meta learning approaches, MAML \cite{finn2017model}, applies two update steps. The first update happens in an inner loop with a meta-train set while the second update happens in outer loop with a meta-test set. In our case, we perform the inner loop update on the discriminator and generator parameters using the meta-train set and the outer loop update is performed on the parameters of generator using a meta-test set. 
\begin{algorithm}[h]
\SetKwData{Left}{left}\SetKwData{This}{this}\SetKwData{Up}{up}
\SetKwFunction{Union}{Union}\SetKwFunction{FindCompress}{FindCompress}
\SetKwInOut{Input}{input}\SetKwInOut{Output}{output}
\Input{Datasets $\mathcal{T}$ and $\mathcal{A}$, model $f$}
\Output{Generator network $\mathcal{G}$ parameters $\boldsymbol{\theta}_g$}
\BlankLine
\While{not converge}{
\For{samples in $\mathcal{T}$}{
\tcc*[h]{Obtain the mask}\\
$\mathbf{M}$ $\leftarrow$ $\sigma$($\lvert \nabla_{\mathbf{x}}{\mathcal{L}_{adv-trip}(\mathbf{x},\mathbf{x}^n,\mathbf{x}^p) } \rvert$)\\
\tcc*[h]{Meta train update using $\mathcal{T}$}\\
 $\boldsymbol{\theta}_d \leftarrow \argmax_{\boldsymbol{\theta}_d} E_{\mathbf{x}}\log \mathcal{D}(\mathbf{x}) + E_{\mathbf{x}}\log(1 - \mathcal{D}(\Pi(\mathcal{G}(\mathbf{x}))))$ \\
 $\boldsymbol{\theta}_g \leftarrow \argmin_{\boldsymbol{\theta}_g}  \mathcal{L}_{\mathcal{G}}^{\mathcal{T}} + \lambda \mathcal{L}_{adv-trip}^{\mathcal{T}}(\mathbf{x}_{adv}^a,\mathbf{x}_{adv}^n,\mathbf{x}_{adv}^p)$\\
 $\boldsymbol{\delta} = \mathbf{x} - \Pi(G(\mathbf{x}))$\\
 \tcc*[h]{Meta test loss using $\mathcal{A}$}\\
  Sample triplets from meta-test set $\mathcal{A}$ and compute $\mathcal{L} = \mathcal{L}_{adv-trip}^{\mathcal{A}}(\mathbf{x}^a - \boldsymbol{\delta},\mathbf{x}^n,\mathbf{x}^p)$\\
 }
 \tcc*[h]{Meta test update}\\
 $\boldsymbol{\theta}_g \leftarrow \argmin_{\boldsymbol{\theta}_g} \lambda \mathcal{L}$\\
 }
\caption{{Training for MeGA}}\label{algo_disjdecomp}
% \EndWhile
\end{algorithm}

More formally, given a network $\mathcal{D}$ parametrized by $\boldsymbol{\theta}_d$ and $\mathcal{G}$ parametrized by $\boldsymbol{\theta}_g$, we  perform the meta-training phase to obtain the parameters $\boldsymbol{\theta}_d$ and $\boldsymbol{\theta}_g$. The update steps are given in Algorithm \ref{algo_disjdecomp}.
We also obtain the adversarial perturbation as, $\boldsymbol{\delta} = \mathbf{x} - \Pi(G(\mathbf{x}))$. 

We then apply the meta-testing update using the additional meta-test dataset ${\mathcal{A}}$. In Algorithm \ref{algo_disjdecomp}, 
% \begin{align}
%     \theta_g \leftarrow \min_{\theta_g}  \mathcal{L}_{\mathcal{G}}^{\mathcal{T}} + \lambda (\mathcal{L}_{adv-trip}^{\mathcal{T}}(x_{adv}^a,x_{adv}^n,x_{adv}^p)\\ + \mathcal{L}_{adv-trip}^{\mathcal{A}}(x^a + \delta,x^n,x^p))
%     \label{eq:meta-train}
% \end{align}
 $\mathcal{L}_{\mathcal{G}}^{\mathcal{T}} = E_{\mathbf{x}}\log(1 - \mathcal{D}(\Pi(\mathcal{G}(\mathbf{x}))))$. We discriminate the datasets using superscripts $\mathcal{T}$ for meta-train set and $\mathcal{A}$ for meta-test set. $\mathcal{L}_{adv-trip}^{\mathcal{A}}$ draws its samples $\mathbf{x}$ from $\mathcal{A}$. At the inference stage, we only use $\mathcal{G}$ to generate the adversarial sample.

\subsection{Training in absence of labels}
 Deep mis-ranking loss can be used \cite{wang2020transferable} when the labels are available for $\mathcal{T}$. In this scenario, we present the case where no labels are available. In the absence of labels and inspired by unsupervised contrastive loss \cite{wang2021understanding}, we generate a positive sample $\mathbf{x}_{adv}^p$ by applying augmentation to the given sample $\mathbf{x}_{adv}^a$. The negative sample $\mathbf{x}_{adv}^n$ is generated using batch hard negative sample strategy, {that is we consider all samples except the augmented version of $\mathbf{x}_{adv}^a$ as negative samples and choose the one which is closest to $\mathbf{x}_{adv}^a$}. We then use {Eq. \ref{eq:adv-triplet}} to obtain the adversarial triplet loss. 
\section{Experimental Results}
\subsection{Implementation Details} We implemented the proposed method in Pytorch framework. The GAN architecture is similar to that of the GAN used in \cite{xiao2018generating, isola2017image}. We use the models from Model Zoo \cite{modelzoo} - OSNet \cite{zhou2019omni}, MLFN \cite{chang2018multi}, HACNN \cite{li2018harmonious}, ResNet-50 and ResNet-50-FC512. We also use AlignedReID \cite{zhang2017alignedreid, AlignedReID}, LightMBN \cite{herzog2021lightweight}, and PCB \cite{sun2018beyond, PCB}.
We use an Adam optimizer with a learning rate = $10^{-5}$, $\beta_1$ = $0.5$ and $\beta_2 = 0.999$ and train the model for 40 epochs. We set $m=1$, {$\lambda = 0.01$}, and $\epsilon = 16$. In order to stabilize GAN training, we apply label flipping with 5\% flipped labels. We first present the ablation for mask and meta learning.

\subsection{Effect of mask $\mathbf{M}$}
 We find that when we use mask for Resnet50 and test for different models like MLFN \cite{chang2018multi} and HACNN \cite{li2018harmonious}, there is a substantial gain in the performance as shown in Table \ref{tab:resnet50_mask}. In terms of R-1 accuracy, introduction of mask gives a boost of 42.10\% and 4.8\% for MLFN and HACNN respectively. This indicates that mask provides better transferability. Further, when we evaluate on Resnet50 itself, there is a minor change in performance which could be because mask is learnt using Resnet50 itself. 
\begin{table}[H]
\caption{Trained on Market-1501 \cite{zheng2015scalable}. Setting Market-1501 $\rightarrow$ Market-1501. $l$ indicates Market-1501 labels are used for training. $\mathbf{M}$ indicates the incorporation of mask. 'Before' indicates accuracy on clean samples.} 
\label{tab:resnet50_mask}
\centering
% \resizebox{\columnwidth}{!}
{
% \begin{tabular}{c| c c c | c c c | c c c | c c c|c c c}
    \begin{tabular}{c|c c | c c |  c c }
\hline
Model &\multicolumn{2}{c|}{Resnet50} &\multicolumn{2}{c|}{MLFN} &\multicolumn{2}{c}{HACNN} \\
& mAP &R-1 &mAP& R-1&mAP &R-1 \\
\hline 
 Before & 70.4& 87.9 & 74.3 &90.1 & 75.6& 90.9\\
$l$ & {0.66}  & {0.41} & 3.95 &3.23   & 32.57& 42.01  \\
{$l+\text{AND}$}&{0.56} & {0.35} & 5.39 & 4.55 & 35.13 &44.20\\
{$l+\text{SAND}$}&\textbf{0.51} & \textbf{0.33} & 6.01 & 4.89 & 37.50 &45.11\\
$l+\mathbf{M}$ &0.69  & 0.50  &\textbf{2.80}   & \textbf{1.87} & \textbf{31.73}  & \textbf{39.99} \\
\hline
\end{tabular}
}
\end{table}

\subsection{Effect of meta learning}
We demonstrate the effect of meta learning in Table \ref{tab:resnet50_meta}. In the case of cross-dataset (Resnet50) as well as cross-dataset cross-model (MLFN) setting, we observe that introduction of meta learning gives a significant performance boost. In terms of R-1 accuracy, there is a boost of 69.87\% and 69.29\% respectively for Resnet50 and MLFN. We further observe that Resnet50 does not have a good transferability towards HACNN. This could be due to two reasons. First, Resnet50 is a basic model compared to other superior PRID models. Second, HACNN is built on Inception units \cite{szegedy2017inception}. 
\begin{table}[H]
\caption{Trained on Market-1501 using MSMT-17 \cite{wei2018person} as meta test set. Setting Market-1501 $\rightarrow$ DukeMTMC-reID \cite{zheng2017unlabeled}. $\mathcal{A}$ indicates incorporation of meta learning.}
\label{tab:resnet50_meta}
\centering
% \resizebox{\columnwidth}{!}
{
% \begin{tabular}{c| c c c | c c c | c c c | c c c|c c c}
    \begin{tabular}{c|c c | c c |  c c }
\hline
{Model} &\multicolumn{2}{c|}{Resnet50} &\multicolumn{2}{c|}{MLFN} &\multicolumn{2}{c}{HACNN} \\
& mAP &R-1 &mAP& R-1&mAP &R-1 \\
\hline 
 Before & 58.9 & 78.3 &  63.2& 81.1 & 63.2&80.1 \\
$l$ & 17.96 & 24.86 & 18.25& 24.10 & \textbf{42.75} &\textbf{58.48} \\
$l+\mathcal{A}$ &\textbf{5.80} & \textbf{7.49} & \textbf{6.15} & \textbf{7.4} & 43.12& 58.97\\
\hline
\end{tabular}
}
\end{table}
\subsection{Adversarial attack performance}
We first present the results for cross-model attack in Table \ref{tab:aligned_source_market}. We use AlignedReID model, Market-1501 \cite{zheng2015scalable} as training set and MSMT-17 \cite{wei2018person} as meta test set. The results are reported for Market-1501 and DukeMTMC-reID \cite{zheng2017unlabeled}. In case of Market-1501, it is clearly evident that the proposed method is able to achieve a strong transferability. We can see that  incorporation of meta test set leads to less than halving the mAP and R-1 results compared to case when only labels are used. For instance, mAP and R-1 of AlignedReID goes down from 7.00\% and 6.38\% to 3.51\% and 2.82\% respectively. This is consistently observed for all three models. Further, the combined usage of mask and meta learning ($l+\mathbf{M}+\mathcal{A}$), denoted as MeGA, achieves best results in cross-model case of PCB and HACNN. The respective R-1 improvements are 10.00\% and 9.10\%. Thus our method is extremely effective in generating adversarial samples. 
\begin{table}[H]
\caption{AlignedReID trained on Market-1501 with MSMT-17 as meta test set. M is Market-1501 and D is DukeMTMC-reID. MeGA denotes $l+\mathbf{M}+\mathcal{A}$.}
\label{tab:aligned_source_market}
\centering
\resizebox{\columnwidth}{!}
{
\begin{tabular}{c|c| c c | c c |c c }
\hline
& {Model} &\multicolumn{2}{c|}{AlignedReID} &\multicolumn{2}{c|}{PCB} &\multicolumn{2}{c}{HACNN} \\
& & mAP &R-1 &mAP& R-1&mAP &R-1 \\
\hline 
M $\rightarrow$ M & Before & 77.56 & 91.18 & 78.54 & 92.87 & 75.6&90.9 \\
\cline{2-8}
% w/label OSNET  &16.62  & 18.91&  39.90 \\\hline
&$l$ & 7.00 & 6.38 & 16.46 & 29.69 & 16.39 & 20.16\\
\cline{2-8}
&$l$ + $\mathbf{M}$  & 6.62& 5.93 & 15.96 & 28.94 & 16.01 & 19.47\\ \cline{2-8}
&$l+\mathcal{A}$ & \textbf{3.51}  & \textbf{2.82}  &  8.07 & 13.86 & 5.44& 5.28 \\
\cline{2-8}
&MeGA& 5.50 & 5.07 &  \textbf{7.39} &\textbf{12.47} & \textbf{4.85} & \textbf{4.80} \\ 
\hline
M $\rightarrow$ D& $l$ & 16.04 & 21.14 & 13.35 & 15.66 & 15.94 & 21.85 \\
\cline{2-8}
&$l+\mathbf{M}$ & 16.23 & 21.72  & 13.70 & 15.97  & 16.43 & 22.17  \\
\cline{2-8}
&$l+\mathcal{A}$ & \textbf{4.69}  & \textbf{5.70}  & \textbf{11.10} & \textbf{12.88} & 5.40 & 6.55\\
\cline{2-8}
&MeGA & 7.70 & 9.47 & 11.81 & 14.04& \textbf{4.73} & \textbf{5.40} \\ 
\hline
\end{tabular}
}
\end{table}

In case of Market-1501 to DukeMTMC-reID, we observe that simply applying the meta learning ($l+\mathcal{A}$) generalizes very well. In case of AlignedReID, mAP and R-1 of 4.60\% and 5.70\% respectively, are significantly lower compared to results obtained via $l$ or $l+\mathbf{M}$ settings. The combined setting of mask and meta learning yields better results for HACNN compared to AlignedReID and PCB. This may be because of the fact that learning of mask is still tied to training set and thus may result in overfitting.  
\iffalse
\begin{table}[H]
% \renewcommand{\arraystretch}{0.9}
\caption{AlignedReID trained on Market with MSMT-17 as meta test set. Results are reported for DukeMTMC-reID. Cross dataset and model setting. }
\label{tab:aligned_test_duke}
\centering
\resizebox{\columnwidth}{!}
{
% \begin{tabular}{c| c c c | c c c | c c c | c c c|c c c}
    \begin{tabular}{c| c c c| c c c|c c c}
\hline
\hline
{Model} &\multicolumn{3}{c|}{AlignedReID} &\multicolumn{3}{c|}{PCB} &\multicolumn{3}{c}{HACNN} \\
& mAP &R-1 &R-10 & mAP &R-1 &R-10& mAP &R-1 &R-10 \\
\hline 
$l+\mathbf{M}$ & 16.23 & 21.72 & 37.79 & 13.70 & 15.97 & 36.13 & 16.43 & 22.17 & 35.77 \\
\hline
$l+\mathcal{A}$ & 4.69  & 5.70  &  12.11 & 11.10 & 12.88& 29.30 & 5.40 & 6.55 & 14.13 \\
\hline
MeGA & 7.70 & 9.47 & 19.16 & 11.81 & 14.04 &31.23 & 4.73 & 5.40 & 11.57  \\ 
\hline
\end{tabular}
}
\end{table}
\fi

In Table \ref{tab:market-msmt_meta_duke} we discuss the results for cross-dataset and cross-model case against more models. Here also we can see that both AlignedReID and PCB lead to strong attacks against other models in a different dataset. 

In Table \ref{tab:aligned_msmt}, we present the results for MSMT-17. Here, the model is trained using AlignedReID and PCB using Market-1501 and DukeMTMC-reID as meta test set. When trained and tested using AlignedReID, the R-1 accuracy drops from 67.6\% on clean samples to 17.69\%. On the other hand when trained using PCB and tested on AlignedReID, the performance drops to 16.70\%. This shows that our attack is very effective in large scale datasets such as MSMT-17.
\tabcolsep=4pt
\begin{table*}[tb]
\caption{AlignedReID and PCB trained on Market with MSMT-17 as meta test set. Setting Market-1501 $\rightarrow$ DukeMTMC-reID.}
\label{tab:market-msmt_meta_duke}
\centering
% \resizebox{\columnwidth}{!}
{
% \begin{tabular}{c| c c c | c c c | c c c | c c c|c c c}
\begin{tabular}{c| c c c |c c c | c c c | c c c | c c c| c c c}
\hline
{Model} &\multicolumn{3}{c|}{OSNet} & \multicolumn{3}{c|}{{LightMBN}} & \multicolumn{3}{c|}{ResNet50} & \multicolumn{3}{c|}{MLFN} & \multicolumn{3}{c|}{ResNet50FC512} & \multicolumn{3}{c}{HACNN} \\
& mAP &R-1 &R-10 & mAP &R-1 &R-10 & mAP &R-1 &R-10 & mAP &R-1 &R-10  & mAP &R-1 &R-10  & mAP &R-1 &R-10 \\
\hline 
Before & 70.2&  87.0 & - & 73.4 & 87.9 & - & 58.9 &  78.3& - & 63.2&  81.1 &-& 64.0 & 81.0& -& 63.2 & 80.1& - \\\hline
AlignedReID & 15.31 & 22.30 & 35.00 & 16.24 & 24.13 &39.65 & 5.17 & 6.64 & 13.77  & 12.28& 16.38 & 29.39 &6.97 & 9.69&19.38 & 4.77 & 5.61 &  11.98\\
\hline
% Before &  &  & - &  &  & - &  &  &-&  &.  & -&   & & - \\
PCB & 12.27  & 14.45 &  27.49 & 12.88 &  15.70 & 28.54& 7.14 &  8.55 & 20.01  & 11.95 &  16.54 &  30.92 & 9.45  & 11.46 & 23.90 & 3.97  & 4.66 &  10.00 \\
\hline
\end{tabular}
}
\end{table*}
\begin{table}[H]
\caption{Trained on Market-1501 using DukeMTMC-reID as meta test set. Setting Market-1501 $\rightarrow$ MSMT-17.}
\label{tab:aligned_msmt}
\centering
% \resizebox{\columnwidth}{!}
{
% \begin{tabular}{c| c c c | c c c | c c c | c c c|c c c}
\begin{tabular}{c| c c c }
\hline
{Model} &\multicolumn{3}{c}{AlignedReID}  \\
& mAP &R-1 &R-10  \\
\hline 
MeGA (AlignedReID)& 9.37 & 17.69 & 33.42 \\ 
\hline
MeGA (PCB) & 8.82 & 16.70 &  31.98\\ 
\hline
\end{tabular}
}
\end{table}
\begin{figure}[H]
\centering
\includegraphics[width = 0.8cm,height = .9cm, cfbox=red 1pt 1pt]{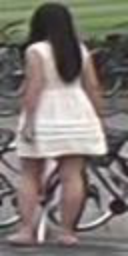} \hspace{4mm} 
\includegraphics[width = 0.8cm,height = .9cm]{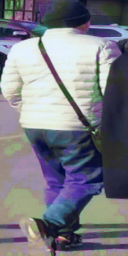}
\includegraphics[width = 0.8cm,height = .9cm]{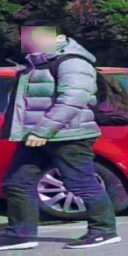}
\includegraphics[width = 0.8cm,height = .9cm]{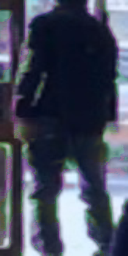}
\includegraphics[width = 0.8cm,height = .9cm]{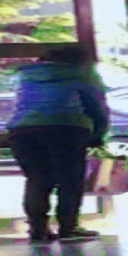}
\includegraphics[width = 0.8cm,height = .9cm]{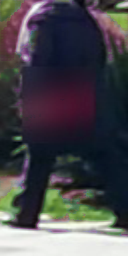}\\
\includegraphics[width = 0.8cm,height = .9cm, cfbox=blue 1pt 1pt]{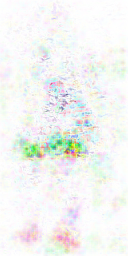} \hspace{4mm}
 \includegraphics[width = 0.8cm,height = .9cm]{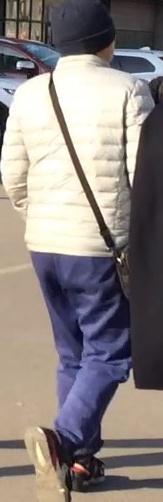}
   \includegraphics[width = 0.8cm,height = .9cm]{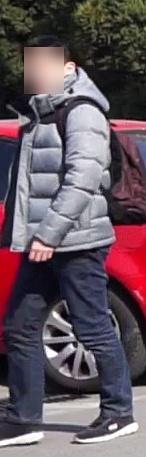}
 \includegraphics[width = 0.8cm,height = .9cm]{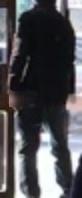}
 \includegraphics[width = 0.8cm,height = .9cm]{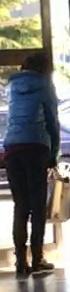}
 \includegraphics[width = 0.8cm,height = .9cm]{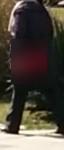}
  \caption{{Left column: Red and blue box show the given image from Market-1501 and its mask ($1-M$) respectively.
  Right column:} Attacked (top) and clean (bottom) images from MSMT-17}
  \label{fig:subjective}
\end{figure}  
\subsection{Comparison with SOTA models}
In Table \ref{tab:comparison_aligned_TCIAA} we present the comparison with TCIAA \cite{wang2020transferable}, UAP \cite{li2019universal} and Meta-attack \cite{yang2021learning}. We observe that our method outperforms TCIAA by a huge margin. We can also see that when mis-ranking loss is naively applied in case of TCIAA$^\dagger$ \cite{yang2021learning}, the model' performance degrades. Our attack has better performance compared to both TCIAA and Meta-attack. 

\begin{table}[H]
\caption{AlignedReID trained on Market with MSMT-17 as meta test set. Setting Market-1501 $\rightarrow$ DukeMTMC-reID. $^\dagger$ uses PersonX \cite{sun2019dissecting} as extra dataset.$^*$ uses PersonX for meta learning. }
\label{tab:comparison_aligned_TCIAA}
\centering
% \resizebox{\columnwidth}{!}
{
% \begin{tabular}{c| c c c | c c c | c c c | c c c|c c c}
\begin{tabular}{c| c c c  }
\hline
{Model} &\multicolumn{3}{c}{Aligned reid}  \\
& mAP &R-1 &R-10  \\
\hline 
Before & 67.81 &80.50 & 93.18 \\
\hline
TCIAA \cite{wang2020transferable}&14.2 & 17.7 & 32.6 \\
{MeGA$^*$ (Ours)} & {11.34} & {12.81} & {24.11} \\ 
MeGA (Ours) & \textbf{7.70} & \textbf{9.47} & \textbf{19.16} \\ 
\hline
 & \multicolumn{3}{c}{PCB} \\
Before & 69.94 &84.47 & - \\
\hline
TCIAA \cite{wang2020transferable} & 31.2 & 45.4 & - \\
TCIAA$^\dagger$ \cite{wang2020transferable} & 38.0 & 51.4 & - \\
UAP \cite{li2019universal} & 29.0 & 41.9 & - \\
Meta-attack$^*$ ($\epsilon = 8$) \cite{yang2021learning} &26.9 & 39.9 & \\
\hline
{MeGA$^*$ ($\epsilon = 8$) (Ours)} & {22.91} & {31.70} & - \\ 
MeGA ($\epsilon = 8$) (Ours) & \textbf{18.01} & \textbf{21.85} & 44.29 \\ 
\hline
\end{tabular}
}
\end{table}

\subsection{Subjective Evaluation}
\begin{figure}[h]
\centering
\includegraphics[width = .9cm,height = 1.1cm, cfbox=blue 1pt 1pt]{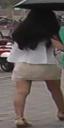}
\includegraphics[width = .9cm,height = 1.1cm, cfbox=green 1pt 1pt]{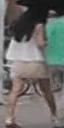}
\includegraphics[width = .9cm,height = 1.1cm, cfbox=green 1pt 1pt]{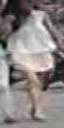}
\includegraphics[width = .9cm,height = 1.1cm, cfbox=green 1pt 1pt]{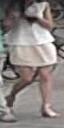}
\includegraphics[width = .9cm,height = 1.1cm, cfbox=red 1pt 1pt]{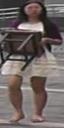}
\includegraphics[width = .9cm,height = 1.1cm, cfbox=red 1pt 1pt]{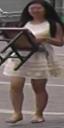}\\
\includegraphics[width = .9cm,height = 1.1cm]{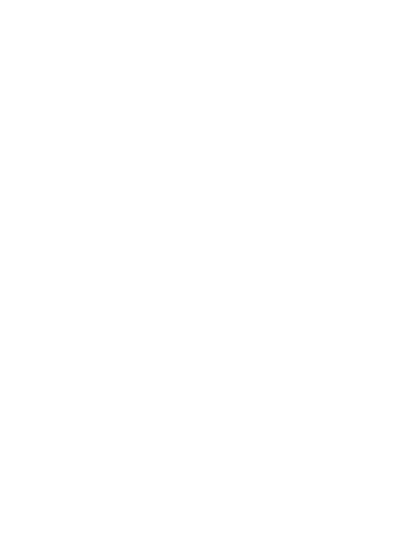}
\includegraphics[width = .9cm,height = 1.1cm, cfbox=red 1pt 1pt]{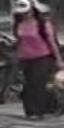}
\includegraphics[width = .9cm,height = 1.1cm, cfbox=red 1pt 1pt]{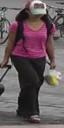}
\includegraphics[width = .9cm,height = 1.1cm, cfbox=red 1pt 1pt]{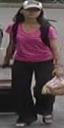}
\includegraphics[width = .9cm,height = 1.1cm, cfbox=red 1pt 1pt]{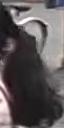}
\includegraphics[width = .9cm,height = 1.1cm, cfbox=red 1pt 1pt]{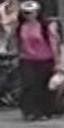}
  \caption{Query image marked in blue border. Top 5 {retrieved} mages from OSNet for Market-1501 (top). Green colored boxes are correct match and red ones are incorrect. Retrieved images after attacking query sample (bottom).}
  \label{fig:retrieved_results}
\end{figure}

We show the example images obtained by our algorithm in Figure \ref{fig:subjective} and top-5 retrieved results in Figure \ref{fig:retrieved_results} for the OSNet model. We can see that in the case of clean samples the top-3 retrieved images match the query ID, however, none of the retrieved images match query ID in the presence of our attack.

\subsection{Attack using unlabelled source}
In this section we discuss the attack when source dataset $\mathcal{T}$ is unlabeled and neither the victim model nor the dataset used for training victim model are available. This is a very challenging scenario as supervised models cannot be used for attack. Towards this, we use unsupervised trained models on Market-1501 and MSMT-17 from \cite{ge2020self}. In Table \ref{tab:train_msmt_test_market}, we present results for training using MSMT-17 and testing on Market. We observe that IBNR50 obtains a mAP and R-1 accuracy of 40.7\% and 52.34\% when both labels and mask are not used. When mask is incorporated there is a substantial boost of 3.82\% in mAP and 4.81\% in R-1 accuracy in case of OSNet. These gains are even higher for MLFN and HACNN. 

In case of Market-1501 to MSMT-17 in Table \ref{tab:market-msmt}, we see that the attack using only mask performs reasonably well compared to that of attacks using label or both label and mask. Due to the comparatively small size of Market-1501, even the attacks using labels are not very efficient.
\begin{table}[H]
\caption{MSMT-17 $\rightarrow$ Market-1501. R50 denotes Resnet50.}
\label{tab:train_msmt_test_market}
\centering
% \resizebox{\columnwidth}{!}
{
% \begin{tabular}{c| c c c | c c c | c c c | c c c|c c c}
\begin{tabular}{c| c c | c c| c c}
\hline
{Model} &\multicolumn{2}{c|}{OSNet} & \multicolumn{2}{c|}{MLFN} & \multicolumn{2}{c}{HACNN} \\
& mAP &R-1 & mAP &R-1  & mAP &R-1  \\
\hline
Before &82.6 & 94.2 & 74.3 & 90.1  & 75.6 & 90. 9 \\
\hline
$l$ (R50) & 30.50 & 39.45   & 26.37 & 38.03  & 31.15 & 39.34\\
$l+\mathbf{M}$ (R50) &24.50 &33.07  & 21.76 & 32.18  & 18.81&23.66  \\
$\mathbf{M}$ (R50) & 36.5 &47.56   & 34.92& 52.61 &31.15 &39.34 \\
% $l$+ OSNet & 12.40 & 14.42  & 14.98 & 21.94  & 10.81 & 11.87 \\
% $l+M$+ OSNet & 12.27& 14.45  & 13.35 &18.94  & 9.00&9.73  \\
\hline
\hline
IBN R50 & 40.7 & 52.34  & 40.62 & 61.46 & 35.44 & 44.84 \\
\hline
$\mathbf{M}$ (IBN R50) & 36.88 & 47.53 & 35.01 & 52.79  & 30.98& 38.98 \\
\hline
\end{tabular}
}
\end{table}
\begin{table}[H]
\caption{ Market-1501 $\rightarrow$ MSMT-17.}
\label{tab:market-msmt}
\centering
% \resizebox{\columnwidth}{!}
{
% \begin{tabular}{c| c c c | c c c | c c c | c c c|c c c}
\begin{tabular}{c| c c | c c| c c}
\hline
{Model} &\multicolumn{2}{c|}{OSNet} & \multicolumn{2}{c|}{MLFN} & \multicolumn{2}{c}{HACNN} \\
& mAP &R-1 & mAP &R-1  & mAP &R-1  \\
\hline 
Before & 43.8 & 74.9 & 37.2 & 66.4 & 37.2 &64.7\\
$l$ (R50) & 31.78 & 60.43  & 25.17 & 49.33 & 28.9&54.91\\
$l+\mathbf{M}$ (R50) &29.04 &56.11  & 22.02 & 43.57 &28.26 &53.53 \\
\hline
$\mathbf{M}$ (R50) & 35.16 & 66.28  &29.16 & 56.65 &29.69& 57.81  \\
\hline
\end{tabular}
}
\end{table}
\section{Conclusion}
We present a generative adversarial attack method using mask and meta-learning techniques. The mask allows better transferability across different networks, whereas, meta learning allows better generalizability. We present elaborate results under various settings. Our ablation also shows the importance of mask and meta-learning. Elaborate experiments on Market-1501, MSMT-17 and DukeMTMC-reID shows the efficacy of the proposed method.

\bibliographystyle{IEEEtran}
\bibliography{IEEEexample}
\end{document}